\pdfoutput=1
\PassOptionsToPackage{table,,pdftex,dvipsnames}{xcolor}

\documentclass[11pt,table,x11names]{article}

\usepackage{acl}

\usepackage{times}
\usepackage{latexsym}

\usepackage[T1]{fontenc}

\usepackage[utf8]{inputenc}

\usepackage{microtype}
\usepackage{balance}
\usepackage{booktabs}

\usepackage{multirow}
\usepackage{dblfloatfix}
\usepackage{graphics}
\usepackage{float}
\restylefloat{table}
\usepackage[table]{xcolor}
\usepackage{adjustbox}
\usepackage{amsmath}
\usepackage{amssymb}
\usepackage{tikz}
\usetikzlibrary{positioning,arrows}
\usepackage{bm}
\usepackage{array,graphicx}
\usepackage[normalem]{ulem}
\usepackage[noabbrev,capitalise]{cleveref}
\definecolor{atomictangerine}{rgb}{1.0, 0.6, 0.4}

\usepackage{xargs}                      
\usepackage[pdftex,dvipsnames]{xcolor}  
\usepackage[colorinlistoftodos,prependcaption,textsize=tiny]{todonotes}
\newcommandx{\czcomm}[2][1=]{\todo[linecolor=red,backgroundcolor=red!25,bordercolor=red,#1]{#2}}

\newcommand\ELBO{$\mathcal{L}_\mathrm{ELBO}$}
\newcommand\INF{$\mathcal{L}_\mathrm{INF}$}
\newcommand\ADV{$\mathcal{L}_\mathrm{ADV}$}
\newcommand\MIN{$\mathcal{L}_\mathrm{MIN}$}

\title{Learning Disentangled Representations of Negation and Uncertainty}

\author{Jake Vasilakes\textsuperscript{1}, Chrysoula Zerva\textsuperscript{2,3}, Makoto Miwa\textsuperscript{4,5}, Sophia Ananiadou\textsuperscript{1,5} \\
  \textsuperscript{1}National Centre for Text Mining, \\
  Department of Computer Science, The University of Manchester \\
  \textsuperscript{2}Instituto Superior Técnico, \textsuperscript{3}Instituto de Telecomunicações \\
  \textsuperscript{4}Toyota Technological Institute \\
 \textsuperscript{5}Artificial Intelligence Research Center, \\National Institute of Advanced Industrial Science and Technology\\
  \texttt{\{jake.vasilakes, sophia.ananiadou\}@manchester.ac.uk},\\
  \texttt{chrysoula.zerva@tecnico.ulisboa.pt} \\
  \texttt{makoto-miwa@toyota-ti.ac.jp} \\
}

\begin{document}
\maketitle

\begin{abstract}
Negation and uncertainty modeling are long-standing tasks in natural language processing. Linguistic theory postulates that expressions of negation and uncertainty are semantically independent from each other and the content they modify. However, previous works on representation learning do not explicitly model this independence. We therefore attempt to disentangle the representations of negation, uncertainty, and content using a Variational Autoencoder\footnote{We make our implementation and data available at \url{https://github.com/jvasilakes/disentanglement-vae}}. We find that simply supervising the latent representations results in good disentanglement, but auxiliary objectives based on adversarial learning and mutual information minimization can provide additional disentanglement gains. 
\end{abstract}

\section{Introduction}
\label{sec:introduction}

In formal semantics, negation and uncertainty are operators whose semantic functions are independent of the propositional content they modify \cite{cann_1993_neg, cann_1993_mod}\footnote{Specifically, the propositional content can be represented by a variable, such as $\neg p$.}. That is, it is possible to form fluent statements by varying only one of these aspects while leaving the others the same. Negation, uncertainty, and content can thus be viewed as {\em disentangled generative factors} of knowledge and belief statements (see \cref{fig:example}).

Disentangled representation learning (DRL) of factors of variation can improve the robustness of representations and their applicability across tasks \cite{Bengio_Courville_Vincent_2013}. Specifically, negation and uncertainty are important for downstream NLP tasks such as sentiment analysis \cite{benamara-etal-2012-negation,Wiegand_Balahur_Roth_Klakow_Montoyo_2010}, question answering \cite{yatskar-2019-qualitative,yang-etal-2016-learning}, and information extraction \cite{Stenetorp_Pyysalo_Ohta_Ananiadou_Tsujii_2012}. Disentangling negation and uncertainty can therefore provide robust representations for these tasks, and disentangling them from content can assist tasks that rely on core content preservation such as controlled generation \cite{Logeswaran_Lee_Bengio_2018} and abstractive summarization \cite{Maynez_Narayan_Bohnet_McDonald_2020}.

\begin{figure}[t!]
    \centering
    \begin{tikzpicture}[mybox/.style={minimum width=4cm,draw,thick,align=center,minimum height=1.8cm}]
\node [draw=gray,fill=atomictangerine!20,thick,rectangle,inner sep=4pt, rounded corners=4pt,text width=5cm,align=center] (X) {  Trees \underline{might} \textbf{not} have leaves.};
\end{tikzpicture}
    \caption{Example indicating the distinction between uncertainty, negation and content. The content ``Trees have leaves'' is modified by the negation (\textbf{bold}) and uncertainty (\underline{underlined}) factors.}
    \label{fig:example}
\end{figure}

Still, no previous work has tested whether negation, uncertainty, and content can be disentangled, as linguistic theory suggests, although previous works have disentangled attributes such as syntax, semantics, and style \cite{balasubramanian_polarized-vae_2021,john-etal-2019-disentangled,cheng-etal-2020-improving,bao-etal-2019-generating,Hu_Yang_Liang_Salakhutdinov_Xing_2017,colombo-etal-2021-novel}. To fill this gap, we aim to answer the following research questions: \\

\noindent
{\bf RQ1:} Is it possible to estimate a model of statements that upholds the proposed statistical independence between negation, uncertainty, and content? \\

\noindent
{\bf RQ2:} A number of existing disentanglement objectives have been explored for text, all giving promising results. How do these objectives compare for enforcing disentanglement on this task? \\

\subsection{Contributions}

In addressing these research questions, we make the following contributions:

\begin{enumerate}
    \item {{\em Generative Model:} We propose a generative model of statements in which negation, uncertainty, and content are independent latent variables. Following previous works, we estimate this model using a Variational Autoencoder (VAE) \cite{Kingma2014AutoEncodingVB,bowman-etal-2016-generating} and compare existing auxiliary objectives for enforcing disentanglement via a suite of evaluation metrics.}
    \item {{\em Simple Latent Representations:} We note that negation and uncertainty have a binary function (positive or negative, certain or uncertain). We therefore attempt to learn corresponding 1-dimensional latent representations for these variables, with a clear separation between each value.}
    \item {{\em Data Augmentation:} Datasets containing negation and uncertainty annotations are relatively small \cite{farkas2010conll,vincze2008bioscope,jimenez2018sfu}, resulting in poor sentence reconstructions according to our preliminary experiments. To address this, we generate weak labels for a large number of Amazon\footnote{\url{https://github.com/fuzhenxin/text_style_transfer}} and Yelp\footnote{\url{https://github.com/shentianxiao/language-style-transfer4}} reviews using a simple na{\"i}ve Bayes classifier with bag-of-words features trained on a smaller dataset of English reviews annotated for negation and uncertainty \cite{konstantinova-etal-2012-review} and use this to estimate our model. Details are given in \cref{sec:data_aug}.}
\end{enumerate}

We note that, in contrast to other works on negation and uncertainty modeling, which focus on token-level tasks of negation and uncertainty cue and scope detection, this work aims to learn \textit{statement}-level representations of our target factors, in line with previous work on text DRL.

\section{Background}

We here provide relevant background on negation and uncertainty processing, disentangled representation learning in NLP, as well as discussion of how this study fits in with previous work.

\subsection{Negation and Uncertainty in NLP}
Negation and uncertainty help determine the asserted veracity of statements and events in text \cite{sauri2009factbank, thompson2017enriching,Kilicoglu_Rosemblat_Rindflesch_2017}, which is crucial for downstream NLP tasks that deal with knowledge and belief. For example, negation detection has been shown to provide strong cues for sentiment analysis \cite{barnes2021improving,ribeiro2020beyond} and uncertainty detection assists with fake news detection \cite{choy2018seeing}. Previous works on negation and uncertainty processing focus on the classification tasks of cue identification and scope detection \cite{farkas2010conll} using sequence models such as conditional random fields (CRFs) \cite{jimenez-zafra-etal-2020-detecting,Li_Lu_2018}, convolutional and recurrent neural networks (CNNs and RNNs) \cite{qian2016speculation,adel2017exploring,ren2018detecting}, LSTMs \cite{Fancellu_Lopez_Webber_2016,Lazib_Zhao_Qin_Liu_2019}, and, most recently, transformer architectures \cite{Khandelwal_Sawant_2020,Lin_Bethard_Dligach_Sadeque_Savova_Miller_2020,zhao-bethard-2020-berts}. While these works focus mostly on learning \textit{local} representations of negation and uncertainty within a sentence, we attempt to learn {\em global} representations that encode high-level information regarding the negation and uncertainty status of statements.

\subsection{Disentangled Representation Learning}
There is currently no agreed-upon definition of disentanglement.
Early works on DRL attempt to learn a single vector space in which each dimension is independent of the others and represents one ground-truth generative factor of the object being modeled \cite{higgins2016beta}.
\citet{higgins_towards_2018} give a group-theoretic definition, according to which generative factors are mapped to independent vector spaces. This definition relaxes the earlier assumption that representations ought to be single-dimensional and formalizes the notion of disentanglement
according to the notion of invariance.
\citet{Shu_Chen_Kumar_Ermon_Poole_2019} decompose the invariance requirement into consistency and restrictiveness, which describe specific ideal properties of the invariances between representations and generative factors.
In addition to independence and invariance, interpretability is an important criterion for disentanglement. \citet{higgins2016beta} point out that while methods such as PCA are able to learn {\em independent} latent representations, because these are not representative of interpretable factors of variation, they are not {\em disentangled}. We therefore want our learned representations to be predictive of meaningful factors of variation. We adopt the term {\em informativeness} from \citet{Eastwood_Williams_2018} to signify this desideratum.

Previous works on DRL for text all use some form of supervision to enforce informativeness of the latent representations.
\citet{Hu_Yang_Liang_Salakhutdinov_Xing_2017}, \citet{john-etal-2019-disentangled}, \citet{cheng-etal-2020-improving}, and \citet{bao-etal-2019-generating} all use gold-standard labels of the generative factors, while other works employ similarity metrics \cite{Chen_Batmanghelich_2020,balasubramanian_polarized-vae_2021}. In contrast, our approach uses weak labels for negation and uncertainty generated using a classifier trained on a small set of gold-standard data. 

These previous works on text DRL all use a similar architecture: a sequence VAE \cite{Kingma2014AutoEncodingVB,bowman-etal-2016-generating} maps inputs to $L$ distinct vector spaces, each of which are constrained to represent a different target generative factor via a supervision signal. We also employ this overall architecture for model estimation and use it as a basis for experimenting with existing disentanglement objectives based on adversarial learning \cite{john-etal-2019-disentangled,bao-etal-2019-generating} and mutual information minimization \cite{cheng-etal-2020-improving}, described in \cref{sec:disentangle}. However, unlike these previous works, which learn high-dimensional representations of all the latent factors, we aim to learn 1-dimensional representations of the negation and uncertainty variables in accordance with their binary function. 

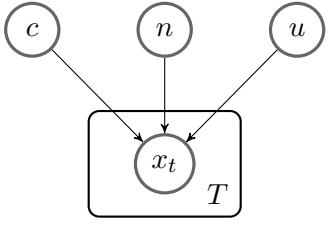
\begin{figure}[t!]
    \centering

    \begin{tikzpicture}[->,>=stealth',
    roundnode/.style={circle, draw=black!60, very thick, minimum size=7mm},
    ]
    \node[roundnode]    (token)                             {$x_t$};
    \node[roundnode]    (negation)      [above=of token]    {$n$};
    \node[roundnode]    (content)       [left=of negation]  {$c$};
    \node[roundnode]    (uncertainty)   [right=of negation]  {$u$};
    
    \draw[black, thick, rounded corners] (-1,0.7) rectangle (1,-0.7);
    \node[] at (0.7,-0.4) {$T$};
    
    \path (content) edge[below,->] node {} (token)
          (negation) edge[below,->] node {} (token)
          (uncertainty) edge[below,->] node {} (token);
    \end{tikzpicture}
    
    \caption{Graph of the generative model. $c$ is content, $n$ is negation status, and $u$ is uncertainty.}
    \label{fig:generative_model}
\end{figure}

\section{Proposed Approach}

We describe our overall model in \cref{sec:overview}. \cref{sec:desiderata} enumerates three specific desiderata for disentangled representations, and sections \ref{sec:informative} and \ref{sec:disentangle}
describe how we aim to satisfy these desiderata.

\subsection{Generative Model}
\label{sec:overview}

We propose a generative model of statements according to which negation,
uncertainty, and content are independent latent variables. A diagram of our proposed model is given in \cref{fig:generative_model}.
Model details are given in \cref{app:model_details}.

We use a sequence VAE to estimate this model \cite{Kingma2014AutoEncodingVB,bowman-etal-2016-generating}. Unlike a standard autoencoder, the VAE imposes a prior distribution on the latent representation space $Z$ (usually a standard Gaussian) and replaces the deterministic encoder with a learned approximation of the posterior $q_{\phi}(z|x)$ parameterized by a neural network.
In addition to minimizing the loss between the input and reconstruction, as in a standard AE, the VAE uses an additional KL divergence term to keep the approximate posterior close to the prior distribution.

\begin{figure}[t!]
    \centering
    \includegraphics[width=0.46\textwidth]{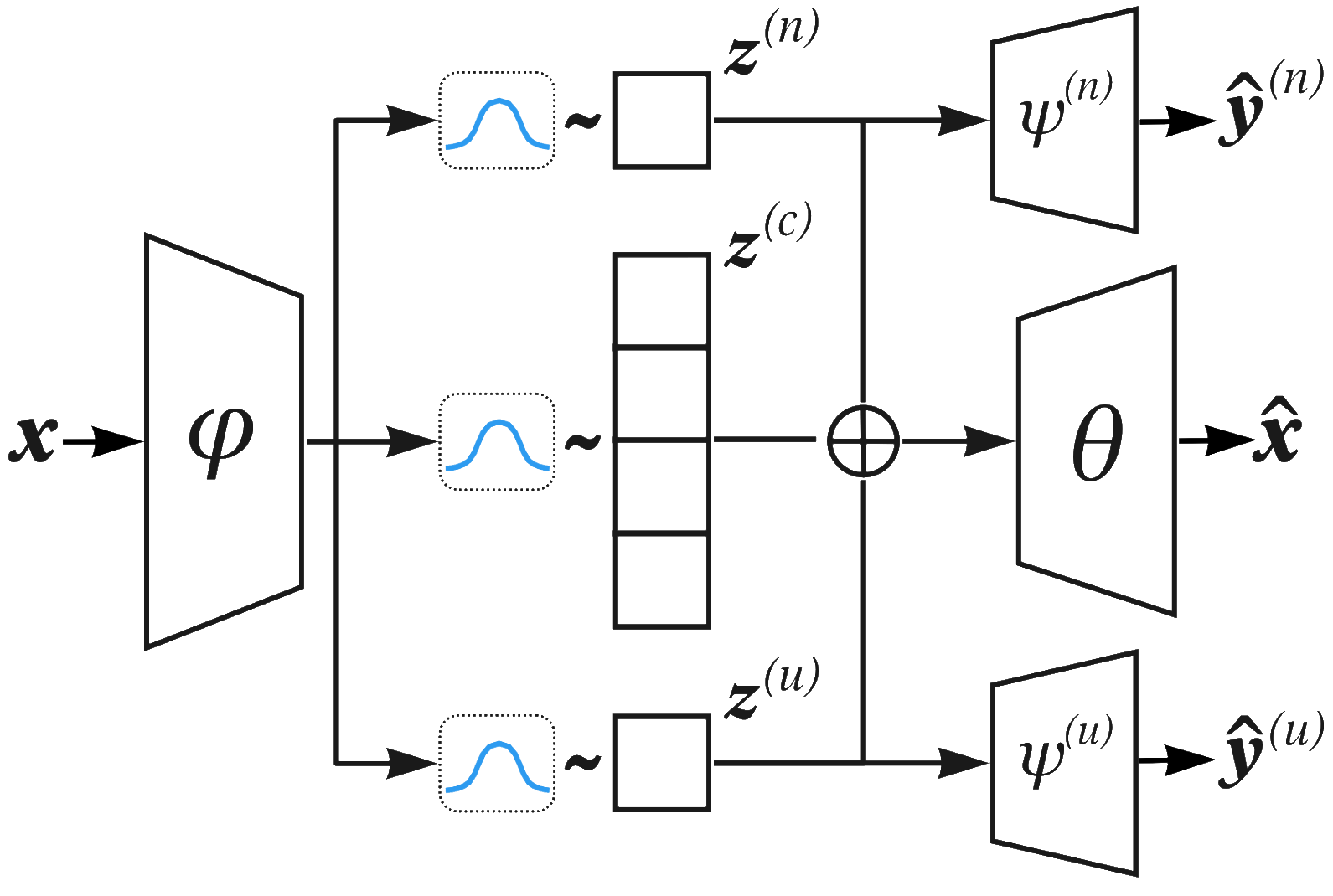}
    \caption{The proposed architecture corresponding to the $\mathcal{L}_{ELBO} + \mathcal{L}_{INF}$ objective (see \cref{sec:disentangle}). A BiLSTM encoder parameterized by $\phi$ maps each input example $x$ to three distinct distributions from which the latent representations $z^{(\ell)}$ are sampled. The negation $z^{(n)}$ and uncertainty $z^{(u)}$ latent spaces are then passed to linear classifiers, parameterized by $\psi^{(\ell)}$, which attempt to predict the ground truth factor. Finally, the latent values initialize an LSTM decoder, parameterized by $\theta$, which attempts to reconstruct the input.}
    \label{fig:figure_arch}
\end{figure}

In our implementation, three linear layers map the final hidden state of a BiLSTM encoder to three sets of Gaussian distribution parameters ($\mu$, $\sigma$), which parameterize the negation, uncertainty, and content latent distributions $\ell \in \{n, u, c\}$, respectively. Because we map each input to three distinct latent spaces, we include three KL divergence terms in the Evidence Lower BOund (ELBO) training objective, given in \cref{eq:elbo_mult}.

\begin{gather}
    \mathcal{L}_{\footnotesize \mathrm{ELBO}}(\phi, \theta) = -\mathbb{E}_{q_{\phi}(z|x)} \Big[ \text{log} ~p_{\theta}(x|z) \Big] \nonumber \\
             + \sum_{\ell \in \{n,u,c\}} \beta_{\ell} ~ \text{KL} \Big[ q^{(\ell)}_{\phi}(z^{(\ell)}|x)~ || ~p(z^{(\ell)}) \Big]
    \label{eq:elbo_mult}
\end{gather}

\noindent
where $\phi$ denotes the encoder's parameters, $\theta$ the decoder's parameters, $p(z^{(\ell)})$ is a standard Gaussian prior, and the $\beta_{\ell}$ hyper-parameters weight the KL divergence term for each latent space $\ell \in L$.
The latent representations $z^{\ell}$ are sampled from normal distributions defined by these parameters using the reparameterization trick \cite{Kingma2014AutoEncodingVB}, i.e., $z^{(\ell)} = \mu^{(\ell)} \odot \sigma^{(\ell)} + \epsilon \sim \mathcal{N}({\bf 0}, {\bf I})$. The latent representations are then concatenated $z = [z^{(n)};z^{(u)};z^{(c)}]$ and used to initialize an LSTM decoder, which aims to reconstruct the input. A visualization of our architecture is given in \cref{fig:figure_arch} and implementation details are given in \cref{app:implementation}.

We use 1-dimensional negation and uncertainty spaces and a 62-dimensional content space for a total latent size of 64. Notably, we do not supervise the content space, unlike previous works \cite{john-etal-2019-disentangled,cheng-etal-2020-improving}, which supervise it by predicting the bag of words of the input. Such a supervision technique would hinder disentanglement by encouraging the content space to be predictive of the negation and uncertainty cues. Therefore, in our model we define three latent spaces $\ell \in \{n, u, c\}$ but use signals from only 2 target generative factors $k \in \{n, u\}$.

\subsection{Desiderata for Disentanglement}
\label{sec:desiderata}

We aim to satisfy the following desiderata of disentangled representations put forth by previous works.

\begin{enumerate}
    \item {\em Informativeness}: the representations should be predictive of the ground-truth generative factors \cite{higgins2016beta,Eastwood_Williams_2018};
    \item {\em Independence}: the representations for each generative factor in question should lie in independent vector spaces \cite{higgins_towards_2018};
    \item {\em Invariance}: the mapping from the data to the representations should be invariant to changes in other generative factors \cite{higgins_towards_2018, Shu_Chen_Kumar_Ermon_Poole_2019};
\end{enumerate}

\noindent
The following sections detail how our model enforces these desiderata.

\subsection{Informativeness}
\label{sec:informative}
Following \citet{Eastwood_Williams_2018}, we measure the informativeness of a representation by its ability to predict the
corresponding generative factor. Similar to previous works on DRL for text \cite{john-etal-2019-disentangled,cheng-etal-2020-improving},
we train supervised linear classifiers\footnote{Implemented as single-layer feed-forward neural networks with sigmoid activation.} on each latent space and back-propagate the prediction error. Thus, in addition to the ELBO objective in \cref{eq:elbo_mult}, we define informativeness objectives for negation and uncertainty.

\begin{equation}
    \mathcal{L}_\mathrm{INF}(\psi^{(k)}) = \mathrm{BCE} \Big( \hat{y}^{(k)}, y^{(k)} \Big), ~~ k \in \{n,u\}
\end{equation}

\noindent
where $y^{(k)}$ is the true label for factor $k$, $\hat{y}^{(k)}$ is the classifier's prediction, $\psi^{(k)}$ are the parameters of this classifier, and BCE is the binary cross-entropy loss.

\subsection{Independence and Invariance}
\label{sec:disentangle}
We compare 3 objectives for enforcing these desiderata:

\begin{enumerate}
    \item Informativeness (INF): This is based on the hypothesis that if negation, uncertainty, and content are independent generative factors, the informativeness objective described in \cref{sec:informative} will be sufficient to drive independence and invariance. This approach was found to yield good results on disentangling style from content by \citet{balasubramanian_polarized-vae_2021}.
   
    \item Adversarial (ADV): The latent representations should be predictive of their target generative factor only. Therefore, inspired by \citet{john-etal-2019-disentangled}, we train additional adversarial classifiers on each latent space that try to predict the values of the non-target generative factors, while the model attempts to structure the latent spaces such that the predictive distribution of these classifiers is a non-predictive as possible. 
   
    \item Mutual-information minimization (MIN): A natural measure of independence between two variables is mutual information (MI). Therefore, this objective minimizes an upper-bound estimate of the MI between each pair of latent spaces, following \cite{Cheng_Hao_Dai_Liu_Gan_Carin_2020,cheng-etal-2020-improving, colombo-etal-2021-novel}.
\end{enumerate}

Details on the ADV and MIN objectives are given below.

\paragraph{Adversarial Objective.}
The adversarial objective (ADV) consists of two parts: 1) adversarial classifiers which attempt to predict the value of all non-target factors from each latent space; 2) a loss that aims to maximize the entropy of the predicted distribution of the adversarial classifiers.

For a given latent space $\ell$, a set of linear classifiers predict the value of each non-target factor $k \ne \ell$, respectively, and we compute the binary cross-entropy loss for each.

\begin{gather}
    \mathcal{L}_\mathrm{CLS}(\xi^{(\ell, k)}) = \mathrm{BCE} \left( \hat{y}^{(\ell, k)}, y^{(k)} \right)
    \label{eq:adv-cls}
\end{gather}

\noindent
Where $\xi^{(\ell, k)}$ are the parameters of the adversarial classifier predicting factor $k$ from latent space $\ell$, and $\hat{y}^{(\ell, k)}$ is the corresponding prediction.

For example, we introduce two such classifiers for the content space $\ell = c$, one to predict negation and one to predict uncertainty, $k \in \{n, u\}$.
Importantly, the prediction errors of these classifiers are not back-propagated to the rest of the VAE. We impose an additional objective for each adversarial classifier, which aims to make it's predicted distribution as close to uniform as possible. We do this by maximizing the entropy of the predicted distribution (\cref{eq:adv-ent}) and back-propagating the error, following \citet{john-etal-2019-disentangled} and \citet{Fu-etal-style}.

\vspace*{-1.5em}
\begin{align}
    \mathcal{L}_\mathrm{ENT}(\xi^{(\ell, k)}) = H[\hat{y}^{(\ell, k)}] 
    \label{eq:adv-ent}
\end{align}

\noindent
As the objective is to maximize this quantity, the total adversarial objective is

\begin{equation}
    \mathcal{L}_\mathrm{ADV} = \sum_{\ell} \sum_k \mathcal{L}_\mathrm{CLS}(\xi^{(\ell, k)}) - \mathcal{L}_\mathrm{ENT}(\xi^{(\ell, k)})
    \label{eq:adv}
\end{equation}

The ADV objective aims to make the latent representations as uninformative as possible for non-target factors. Together with the informativeness objective, it pushes the representations to specialize in their target generative factors.

\paragraph{MI Minimization Objective.}
The MI minimization (MIN) objective focuses on making the distributions of each latent space as dissimilar as possible. We minimize the MI between each pair of latent spaces according to \cref{eq:mi-min}.

\begin{equation}
    \mathcal{L}_\mathrm{MIN} = \hat{I}_\mathrm{CLUB}(\ell_i; \ell_j), ~~i \ne j
    \label{eq:mi-min}
\end{equation}

\noindent
where $\hat{I}_\mathrm{CLUB}(\ell_i; \ell_j)$ is the Contrastive Learning Upper-Bound (CLUB) estimate of the MI \cite{Cheng_Hao_Dai_Liu_Gan_Carin_2020}. Specifically, we introduce a separate neural network to approximate the conditional variational distribution $p_{\sigma}(\ell_i|\ell_j)$, which is used to estimate an upper bound on the MI using samples from the latent spaces.

The full model objective along with relevant hyperparameters weights $\bm{\lambda}$ is given in \cref{eq:full-objective}. Our hyperparameter settings and further implementation details are given in \cref{app:implementation}.

\vspace*{-1.2em}
\begin{align}
    \mathcal{L} = & \mathcal{L}_\mathrm{ELBO} + \lambda_\mathrm{INF}\mathcal{L}_\mathrm{INF} + \nonumber \\  
                         & \lambda_\mathrm{ADV}\mathcal{L}_\mathrm{ADV} + \lambda_\mathrm{MIN}\mathcal{L}_\mathrm{MIN}
    \label{eq:full-objective}
\end{align}

In the sections that follow, we experiment with different subsets of the terms in the full objective and their effects on disentanglement.
We train a model using only the ELBO objective as our disentanglement baseline.

\section{Experiments}

We describe our datasets, preprocessing, and data augmentation methods in \cref{sec:data}. \cref{sec:evaluation} describes our evaluation metrics and how these target the desiderata for disentanglement given in \cref{sec:desiderata}.

\subsection{Datasets}
\label{sec:data}
We use the SFU Review Corpus \cite{konstantinova-etal-2012-review} as our primary dataset. This corpus contains 17,000 sentences from reviews of various products in English, originally intended for sentiment analysis, annotated with negation and uncertainty cues and their scopes. Many of the SFU sentences are quite long ($> 30$ tokens), and preliminary experiments revealed that this results in poor reconstructions. We therefore took advantage of SFU's annotated statement conjunction tokens to split the multi-statement sentences into single-statement ones in order to reduce the complexity and increase the number of examples. Also to reduce complexity, we remove sentences $>15$ tokens following previous work \cite{Hu_Yang_Liang_Salakhutdinov_Xing_2017}, resulting in 14,000 sentences.

We convert all cue-scope annotations to statement-level annotations. Multi-level uncertainty annotations have been shown to be rather inconsistent and noisy, achieving low inter-annotator agreement compared to binary ones \cite{rubin-2007-stating}. We therefore binarize the certainty labels following \cite{zerva2019automatic}.

\subsubsection{Data Augmentation}
\label{sec:data_aug}

Despite the efforts above, we found the SFU corpus alone was insufficient for obtaining fluent reconstructions. We therefore generated weak negation and uncertainty labels for a large amount of additional Amazon and Yelp review data using two na{\"i}ve Bayes classifiers with bag-of-words (BOW) features\footnote{Implementation details and evaluation of these classifiers is given in \cref{app:bow}}. These classifiers were trained on the SFU training split to predict sentence level negation and uncertainty, respectively.
The Amazon and Yelp datasets fit the SFU data distribution well, being also comprised of user reviews, and have been used in previous works on text DRL with good results \cite{john-etal-2019-disentangled,cheng-etal-2020-improving}\footnote{Due to computational constraints, we randomly sample 100,000 weakly annotated Amazon examples for the final dataset. Preliminary experiments with larger numbers of Amazon examples suggested that 100k is sufficient for our purposes.}. Statistics for the combined SFU+Amazon dataset are summarized in \cref{app:amazon_results}. In \cref{app:yelp_results}, we provide a complementary evaluation on a combined SFU+Yelp dataset.

\subsection{Evaluation}
\label{sec:evaluation}

Evaluating disentanglement of the learned representations requires complementary metrics of the desiderata given in \cref{sec:desiderata}: informativeness, independence, and invariance. 

For measuring informativeness, we report the precision, recall, and F1 score of a logistic regression model trained to predict each of the ground-truth labels from each latent space, following \citet{Eastwood_Williams_2018}. We also report the MI between each latent distribution and factor, as this gives additional insight into informativeness.

\begin{figure}[t!]
    \centering
    \includegraphics[width=0.47\textwidth]{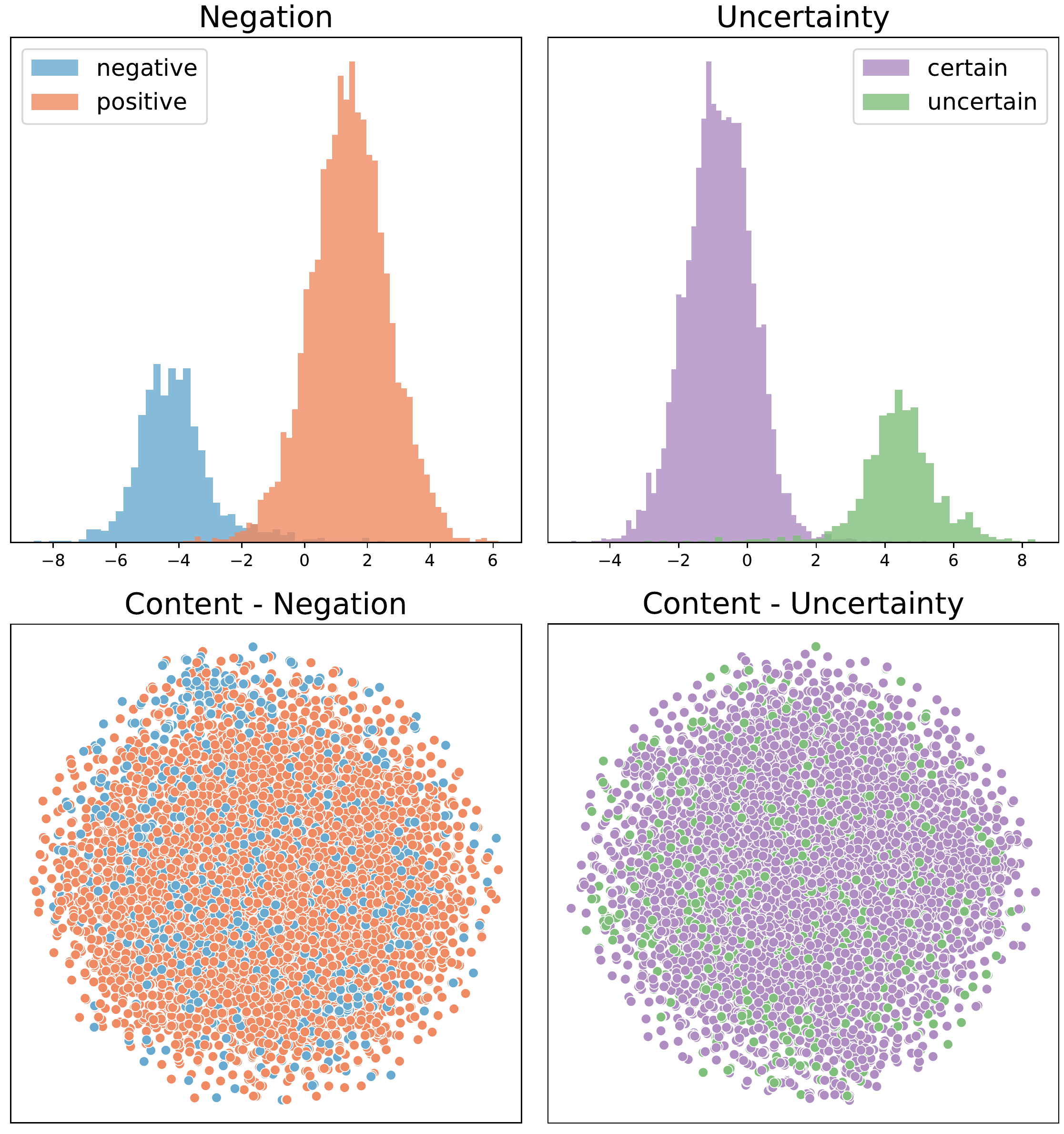}
    \caption{Histogram and t-SNE \cite{van2008visualizing} visualizations of the modality, negation, and content spaces learned by the INF+ADV+MIN model.}
    \label{fig:latent_viz}
\end{figure}

For measuring independence, we use the Mutual Information Gap (MIG) \cite{Chen_Li_Grosse_Duvenaud_2018}. The MIG lies in $[0, 1]$, with higher values indicating a greater degree of disentanglement. Details of the MIG computation are give in \cref{app:mig}.

We evaluate invariance by computing the Pearson's correlation coefficient between each pair of latent variables using samples from the predicted latent distributions.

\begin{table*}[ht!]
\resizebox{\textwidth}{!}{%
    \centering
    \begin{tabular}{cc||cccc|cccc|cccc}
                              &        &  \multicolumn{4}{c|}{\ELBO}     &     \multicolumn{4}{c|}{+\INF}     & \multicolumn{4}{c}{+\INF+\ADV+\MIN}  \\
        Latent                & Factor &  MI    &  P    & R     & F1     &      MI   &  P    & R     & F1    &  MI    &  P    & R     & F1       \\
        \hline                                     
        \multirow{2}{*}{$n$}  &  $n$   &  0.018 & 0.561 & 0.615 & 0.530  &   0.434 & 0.951 & 0.972 & 0.961   & 0.444 & 0.959 & 0.975 & 0.967    \\
                              &  $u$   &  0.016 & 0.558 & 0.590 & 0.533  &   0.013 & 0.551 & 0.584 & 0.544   & 0.007 & 0.543 & 0.569 & 0.537    \\
        \hline
        \multirow{2}{*}{$u$}  &  $n$   &  0.011 & 0.559 & 0.606 & 0.540  &   0.013 & 0.555 & 0.579 & 0.550   & 0.007 & 0.548 & 0.560 & 0.548    \\
                              &  $u$   &  0.022 & 0.570 & 0.604 & 0.562  &   0.375 & 0.936 & 0.972 & 0.952   & 0.391 & 0.970 & 0.981 & 0.975    \\
        \hline
        \multirow{2}{*}{$c$}  &  $n$   &  0.297 & 0.683 & 0.760 & 0.695  &   0.222 & 0.675 & 0.753 & 0.684   & 0.147 & 0.576 & 0.617 & 0.557    \\
                              &  $u$   &  0.207 & 0.653 & 0.756 & 0.665  &   0.198 & 0.643 & 0.746 & 0.649   & 0.136 & 0.574 & 0.637 & 0.551    \\
        \hline
    \end{tabular}
    } 
    \caption{Mutual Information estimate between each latent and factor. Precision, recall, and F1 of the latent space classifiers for each factor. $n$: negation. $u$: uncertainty. $c$: content. Shown are the mean values computed from 30 resamples of the latent distributions for each example on the SFU+Amazon test set. Because their results are similar to \ELBO+\INF+\ADV+\MIN, the \ELBO+\{\ADV,\MIN\} are not included to save space. We provide a more extensive set of results covering all models in \cref{tab:prediction_results_plus}.}
    \label{tab:prediction_results}
\end{table*}

It also important to evaluate the ability of the models to reconstruct the input. Specifically, we target reconstruction faithfulness (i.e., how well the input and reconstruction match) and fluency. We evaluate faithfulness in terms of the ability of the model to preserve the negation, uncertainty, and content of the input. Negation and uncertainty preservation are measured by re-encoding the reconstructions, predicting the negation and uncertainty statuses from the re-encoded latent values, and computing precision, recall, and F1 score against the ground-truth labels\footnote{This corresponds to the measure of consistency proposed by \citet{Shu_Chen_Kumar_Ermon_Poole_2019}}. Following previous work, we approximate a measure of content preservation in the absence of any explicit content annotations by computing the BLEU score between the input and the reconstruction (self-BLEU) \cite{bao-etal-2019-generating,cheng-etal-2020-improving,balasubramanian_polarized-vae_2021}. We evaluate fluency of the reconstruction by computing the perplexities (PPL) under GPT-2, a strong, general-domain language model \cite{radford2019language}. 

Finally, we evaluate the models' ability to flip the negation or uncertainty status of the input. For each test example, we override the value of the latent factor we want to change to represent the \textit{opposite} of its ground-truth label. The ability of the model to control negation and uncertainty is measured by re-encoding the reconstructions obtained from the overridden latents, predicting from the re-encoded latent values, and computing accuracy against the \textit{opposite} of the ground-truth labels.

\begin{figure*}[ht!]
    \centering
    \includegraphics[width=0.95\textwidth]{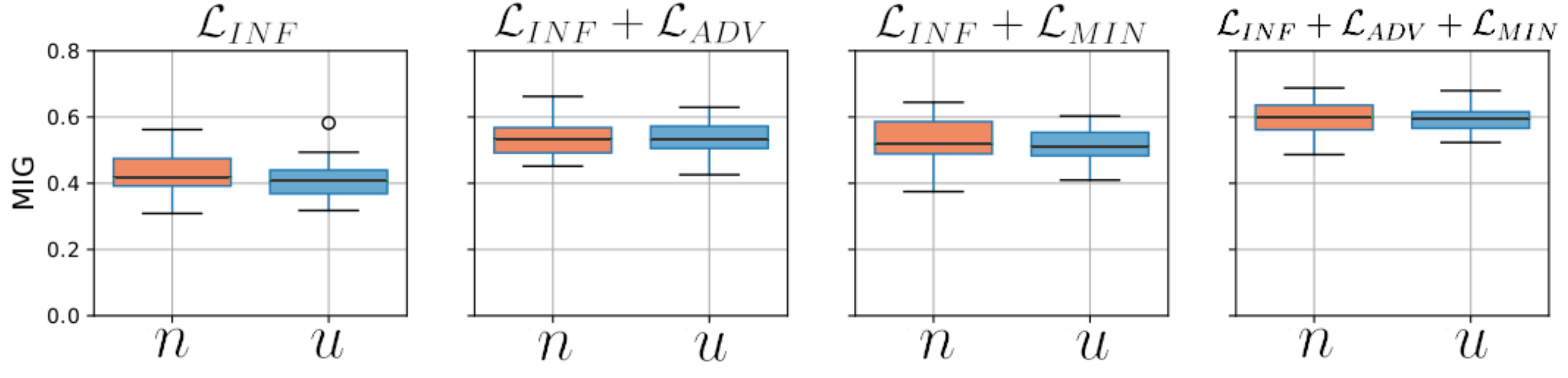}
    \caption{Box plots of the Mutual Information Gap (MIG) for each disentanglement objective for the negation and uncertainty factors computed on the test set. The MIG values for the baseline \ELBO  ~objective were too small to include in this figure, at $\approx 0.014$ for both negation and uncertainty.}
    \label{fig:mig}
\end{figure*}

\section{Results}

In the following, \cref{sec:disentangle_results} reports the disentanglement results and \cref{sec:fluency_results} reports the faithfulness and fluency results. \cref{sec:discussion} discusses how these results address the two research questions proposed in \cref{sec:introduction}.

\subsection{Disentanglement}
\label{sec:disentangle_results}

The informativeness of each latent space with respect to each target factor is shown in \cref{tab:prediction_results} given as predictive performance and MI.

The baseline ELBO objective alone fails to disentangle. It puts almost all representation power in the content space, which is nevertheless still uninformative of the negation and uncertainty factors, with low MIs and F1s. The model using the INF auxiliary objective does, however, manage to achieve good disentanglement: the negation and uncertainty spaces are highly informative of their target factors and uninformative of the non-target factors\footnote{Experiments using \ELBO+\ADV ~or \ELBO+\MIN ~did not improve over \ELBO ~alone.}. However, the content space is still slightly predictive of negation and uncertainty, with F1s of 0.684 and 0.649, respectively. This improves with the ADV and MIN objectives, where the content space shows near-random prediction performance of negation and uncertainty, with slightly improved prediction performance of the negation and uncertainty spaces for their target factors. These results are corroborated by the visualizations in \cref{fig:latent_viz}, which show clear separation by classes in the negation and uncertainty latent distributions but no distinction between classes in the content space. Additionally, we note the good predictive performance of the negation and uncertainty latents, despite their simple, 1-dimensional encoding.

\newcommand\mrt[1]{\multirow{2}{*}{#1}}
\begin{table}[h!]
\small
    \centering
    \begin{adjustbox}{center}
    \begin{tabular}{c|cc|cc|cc}
                                \multicolumn{3}{c}{}           & \multicolumn{2}{c}{}         &  \multicolumn{2}{c}{+\INF+\ADV}     \\
        \multicolumn{1}{c}{}  & \multicolumn{2}{c}{\ELBO}      & \multicolumn{2}{c}{+\INF}    &  \multicolumn{2}{c}{+\MIN}         \\
                              & $u$        & $c$               & $u$        & $c$             & $u$       & $c$                   \\
        \hline
         \mrt{$n$}            & \mrt{0.706} & 0.008            & \mrt{0.200} & 0.002          & \mrt{0.159}     & 0.001           \\
                              &       & {\tiny ($\pm0.053$)}   &       & {\tiny ($\pm0.098$)} &           & {\tiny ($\pm0.043$)}  \\
        \hline
         \mrt{$u$}            & \mrt{-} & 0.001                & \mrt{-}     & 0.001          & \mrt{-}       & 0.005                 \\
                              &     & {\tiny ($\pm0.058$)}     &     & {\tiny ($\pm0.097$)}   &     & {\tiny ($\pm0.037$)}  \\
        \hline
    \end{tabular}
    \end{adjustbox}
    \caption{Pearson's correlation coefficients between each pair of latent representations across models. As the content space is 62-dimensional, we compute the correlation coefficient between each dimension and report the mean with the standard deviation in parentheses below.}
    \label{tab:covariances}
\end{table}

Box plots of the MIG values for the negation and uncertainty factors are given in \cref{fig:mig}. Again we see that the INF objective alone results in decent disentanglement, with median MIG values $\approx0.4$. The ADV and MI objectives give similar increases in MIG, up to $\approx0.55$ for both negation and uncertainty, and their combination, ADV+MIN, improves MIG further, up to $\approx0.6$, suggesting that these objectives are complementary.

We demonstrate the invariance of our models' negation and uncertainty representations in \cref{tab:covariances}. While the ELBO objective alone results in highly covariant negation and uncertainty latent distributions (0.706), this drops significantly under INF (0.200) with additional reduction contributed by the ADV and MIN objectives (0.159).

\vspace*{-0.5em}
\subsection{Evaluation of Reconstructions}
\label{sec:fluency_results}

\subsubsection{Faithfulness and Fluency}
\cref{tab:generation_results} reports the self-BLEU and perplexity for each disentanglement objective. Example reconstructions are given in \cref{app:reconstructions_amazon}. These results show that the models are quite consistent regarding content reconstruction on the train set, but this consistency drops on dev and test. While the ADV and MIN objectives provide disentanglement gains over INF, the BLEU scores betray a possible trade off of slightly poorer content preservation, despite better perplexities.

\begin{table}[t!]
\resizebox{0.5\textwidth}{!}{%
    \centering
    \begin{tabular}{cl|ccccc}
        \multicolumn{2}{c}{}          &       &      &        &       & +\INF      \\
        \multicolumn{2}{c}{}          &       &      & +\INF  & +\INF & +\ADV      \\
                              &       & \ELBO &+\INF & +\ADV  & +\MIN & +\MIN      \\
        \hline
        \parbox[t]{0mm}{\multirow{3}{*}{\rotatebox[origin=c]{90}{BLEU}}} 
                              & Train & 0.590 & 0.576 & 0.574 & 0.528 &  0.522    \\
                              & Dev   & 0.150 & 0.189 & 0.186 & 0.146 &  0.141    \\
                              & Test  & 0.153 & 0.072 & 0.073 & 0.148 &  0.144    \\
        \hline\hline
        \parbox[t]{0mm}{\multirow{3}{*}{\rotatebox[origin=c]{90}{PPL}}}
                             & Train  & 123.3 & 174.3 & 173.1 & 124.9 & 127.2      \\
                             & Dev    & 136.4 & 186.1 & 189.1 & 140.1 & 141.3      \\
                             & Test   & 136.8 & 185.9 & 187.3 & 139.3 & 142.1      \\
        \hline
    \end{tabular}
    } 
    \caption{Reconstruction self-BLEU and reconstruction perplexity (PPL) for each model on each data split. Perplexity is computed using GPT-2 \cite{radford2019language}.}
    \label{tab:generation_results}
\end{table}

While self-BLEU indicates the consistency of the reconstructions with respect to content, it does not necessarily indicate consistency of the reconstructions with respect to negation and uncertainty, which often differ from their opposite value counterparts by a single token. The consistency of the INF and INF+ADV+MIN models with respect to these factors is reported in \cref{tab:consistency_results}. The INF objective alone is only somewhat consistent, with re-encoded F1s of 0.830 and 0.789 for negation and uncertainty respectively. The auxiliary objectives improve these considerably, to 0.914 and 0.893, suggesting that the disentanglement gains seen in \cref{tab:prediction_results} and \cref{fig:mig} have a positive effect on the consistency of the reconstructions.

\begin{table}[t!]
    \centering
    \begin{adjustbox}{center}
    \resizebox{0.5\textwidth}{!}{%
    \begin{tabular}{lc|ccc|ccc}
        \multicolumn{2}{c}{} & \multicolumn{3}{c}{}     & \multicolumn{3}{c}{+\INF+\ADV}  \\
        \multicolumn{2}{c}{} & \multicolumn{3}{c}{+\INF}& \multicolumn{3}{c}{+\MIN}  \\
        Factor       &  Pass & P       & R      & F1    & P       & R      & F1          \\
        \hline
\multirow{2}{*}{$n$} & 1     & 0.969   & 0.965  & 0.967 & 0.959   & 0.975  & 0.967 \\
                     & 2     & 0.816   & 0.848  & 0.830 & 0.920   & 0.908  & 0.914 \\
        \hline
\multirow{2}{*}{$u$} & 1     & 0.959   & 0.961  & 0.960 & 0.970   & 0.981  & 0.975 \\
                     & 2     & 0.767   & 0.820  & 0.789 & 0.930   & 0.864  & 0.893 \\
        \hline
    \end{tabular}
    }
    \end{adjustbox}
    \caption{Consistency of the decoder with the ground-truth values of negation and uncertainty evaluated on the test set. Pass 1 refers to the predictions from the original inputs. Pass 2 refers to the predictions from the re-encoded reconstructions. Pass 1 can be considered an upper bound on the performance of pass 2.}
    \label{tab:consistency_results}
\end{table}

\subsubsection{Controlled Generation}
\label{sec:controlled_generation}

\cref{tab:control} shows the accuracies of each model on the controlled generation task, split by transfer direction. We found that for both negation and uncertainty modifying the status of the input works well in only one direction: from negated to positive, uncertain to certain. 

Changing a sentence from negated to positive or from uncertain to certain generally requires the \textit{removal} of cue tokens (e.g., \textit{not, never, might}), while the opposite directions require their \textit{addition}. Via linear regressions between the content representations and number of tokens, we found that the content space is highly informative of sentence length, which effectively bars the decoder from adding the required negation or uncertainty tokens\footnote{The tendency of VAEs to focus their representation on sentence length was also observed by \citet{Bosc_Vincent_2020}.}. A manual review of correctly and incorrectly modified sentences suggested that the decoder attempts to represent the negation/uncertainty status by \textit{modifying} tokens in the input, rather than adding or removing them, in order to satisfy the length constraint. When \textit{removal} is required, the cue token is often simply replaced by new tokens consistent with the representation. The inclusion of negation/uncertainty cue tokens, however, only seems to occur when it is possible to change an existing token to a cue token. Details of the linear regressions as well as example successful/failed transfers are given in \cref{app:control_results_amazon}.

\begin{table}[h]
    \centering
    \begin{adjustbox}{center}
    \resizebox{0.5\textwidth}{!}{%
    \begin{tabular}{c|cccc}
        \multicolumn{2}{c}{}          &       &       & +\INF \\
        \multicolumn{2}{c}{}          & +\INF & +\INF & +\ADV \\
        Transfer direction    & +\INF & +\ADV & +\MIN & +\MIN \\
        \hline
        pos $\rightarrow$ neg & 0.30  & 0.50  & 0.35  & 0.38  \\
        neg $\rightarrow$ pos & 0.80  & 0.92  & 0.79  & 0.87  \\
        \hline
        cer $\rightarrow$ unc & 0.34  & 0.40  & 0.32  & 0.36  \\
        unc $\rightarrow$ cer & 0.85  & 0.88  & 0.80  & 0.86  \\
        \hline
    \end{tabular}
    }
    \end{adjustbox}
    \caption{Controlled generation accuracies by factor and direction of transfer.}
    \label{tab:control}
\end{table}

\subsection{Research Questions}
\label{sec:discussion}

{\bf RQ1:} {\em Is it possible to learn disentangled representations of negation, uncertainty, and content?} \\
The results suggest that it is indeed possible to estimate a statistical model in which negation, uncertainty, and content are disentangled latent variables according to our three desiderata outlined in \cref{sec:desiderata}. Specifically, \cref{tab:prediction_results} shows high informativeness of the negation and uncertainty spaces across objectives, and the poor predictive ability of each latent space for non-target factors suggests independence. \cref{fig:mig} further suggests independence across models, with median MIG scores in the 0.4-0.6 range. Finally, the low covariances in \cref{tab:covariances} demonstrates the invariance of the latent representations to each other. \\

\noindent
{\bf RQ2:} {\em How do existing disentanglement objectives compare for this task?}\\
Notably, the INF objective alone results in good disentanglement according to our three desiderata, suggesting that supervision alone is sufficient for disentanglement.
Still, the addition of the ADV and MIN objectives resulted in slightly more informative (\cref{tab:prediction_results}) and independent (\cref{tab:covariances}) representations. While the self-BLEU scores reported in \cref{tab:generation_results} suggest that content preservation is generally maintained across auxiliary objectives, small dips are seen in those using the MIN objective. This trend also holds for perplexity, suggesting that while the MIN objective can contribute to disentanglement gains, it may result in poorer reconstructions.

\section{Conclusion}
Motivated by linguistic theory, we proposed a generative model of statements in which negation, uncertainty, and content are disentangled latent variables. We estimated this model using a VAE, comparing the performance of existing disentanglement objectives. Via a suite of evaluations, we showed that it is indeed possible to disentangle these factors. While objectives based on adversarial learning and MI minimization resulted in disentanglement and consistency gains, we found that a decent balance between variable disentanglement and reconstruction ability was obtained by a simple supervision of the latent representations (i.e., the INF objective). Also, our 1-dimensional negation and uncertainty representations achieved high predictive performance, despite their simplicity. Future work will explore alternative latent distributions, such as discrete distributions \cite{Jang_Gu_Poole_2017,Dupont_2018}, which may better represent these operators.

This work has some limitations. First, our model does not handle negation and uncertainty scope, but rather assumes that operators scope over the entire statement. Our model was estimated on relatively short, single-statement sentences to satisfy this assumption, but future work will investigate how operator disentanglement can be unified with models of operator scope in order to apply it to longer examples with multiple clauses. Second, while our models achieved high disentanglement, they fell short on the controlled generation task. We found that this was likely due to the models memorizing sentence length, constraining the reconstructions in way that is incompatible with the addition of negation and uncertainty cue tokens. \citep{Bosc_Vincent_2020} also noticed this tendency for sentence length memorization in VAEs and future will will explore their suggested remedies, such as encoder pretraining.

\section*{Acknowledgements}
This paper is based on results obtained from a project, JPNP20006, commissioned by the New Energy and Industrial Technology Development Organization (NEDO). This work was also supported by the University of Manchester President's Doctoral Scholar award, a collaboration between the University of Manchester and the Artificial Intelligence Research Center, the European Research Council (ERC StG DeepSPIN 758969), and by the Fundação para a Ciência e Tecnologia through contract UIDB/50008/2020.

\bibliography{anthology,custom}
\bibliographystyle{acl_natbib}

\appendix

\section{Model Details}
\label{app:model_details}

We here define our generative model and derive the corresponding ELBO objective for our proposed VAE with three latent variables.
Let $x = [x_1, ..., x_T]$ be a sentence with $T$ tokens. $N$, $U$, and $C$ are the latent variables representing negation, uncertainty, and content respectively. The joint probability of specific values of these variables ($N=n$, $U=u$, $C=c$) is defined as

\vspace*{-1em}
\begin{equation}
    p(x,n,u,c) = p_{\theta}(x|n,u,c)p(n)p(u)p(c)
\end{equation}

\noindent
Furthermore, $x$ is defined auto-regressively as 

\vspace*{-1em}
\begin{equation}
    p(x|n,u,c) = \prod_t p(x_t|x_{<t},n,u,c)
\end{equation}

Our model assumes that the latent factors are independent, so the posterior is

\vspace*{-1em}
\begin{align}
    p(n,u,c|x) & = p(n|x)p(u|x)p(c|x) \\
                        & = \prod_{\ell \in \{n,u,c\}}\frac{p(x|\ell)p(\ell)}{p(x)}
\end{align}

We approximate the posterior $p(n,u,c|x)$ with $q_{\phi}(n,u,c|x)$. Because the posterior factors,
we approximate the individual posteriors $p(\cdot|x)$ with $q_{\phi}(\cdot|x)$ and derive the following ELBO objective.

\begin{align}
    \text{ELBO} & = \mathbb{E}_{q_{\phi}(n,u,c|x)} \left[ \text{log} \frac{p(x,n,u,c)}{q_{\phi}(n,u,c|x)} \right] \\
                & = \mathbb{E}_{q_{\phi}} \left[ \text{log} \frac{p(x|n,u,c)p(n)p(u)p(c)}{q_{\phi}(n|x)q_{\phi}(u|x)q_{\phi}(c|x)} \right] \\
                & = \mathbb{E}_{q_{\phi}} \left[ \text{log} ~p(x|n,u,c) \right] \\
                & ~~~~~~ - \sum_{\ell} KL\left[ q_{\phi}(\ell|x) || p(\ell) \right] \nonumber
\end{align}

\section{Bag-of-Words Classifiers}
\label{app:bow}

We here provide the implementation details and evaluation of our bag-of-words (BOW) classifiers used to generate weak labels for the Amazon and Yelp data.

\subsection{Implementation Details}

Both classifiers are implemented as Bernoulli na{\"i}ve Bayes classifiers with BOW features. We used the {\tt BernoulliNB} implementation from scikit-learn with the default parameters in version 0.24.1\citep{scikit-learn}.

\subsubsection{Feature Selection}

We performed feature selection by computing the $K$ tokens from the SFU training data that had the highest ANOVA F-value against the target labels, implemented using {\tt f\_classif} in scikit-learn \citep{scikit-learn}. We tuned $K$ according to the downstream classification performance on the SFU dev set. We evaluated $K$ in the range 3-30 and found $K=20$ performed best for both models. The 20 tokens ultimately used as features by the negation and uncertainty classifiers are given in \cref{tab:bow_features}.

\newcolumntype{L}{>{\centering\arraybackslash}m{5cm}}
\begin{table}[h!]
    \small
    \centering
    \begin{tabular}{|c|L|}
        \hline
        Negation    &  {\em any, but, ca, cannot, did, do, does, dont, either, even, have, i, it, n't, need, never, no, not, without, wo} \\
        \hline
        Uncertainty & {\em be, can, could, either, i, if, may, maybe, might, must, or, perhaps, probably, seem, seemed, seems, should, think, would, you} \\
        \hline
    \end{tabular}
    \caption{The tokens used as features for the negation and uncertainty classifiers. Contractions are split in our tokenization scheme.}
    \label{tab:bow_features}
\end{table}

\subsection{Evaluation}

We report the precision, recall, and F1 score of both classifiers on the SFU dev and test sets in \cref{tab:bow_eval}.

\begin{table}[h!]
    \centering
    \small
    \begin{tabular}{cc|ccc}
                                     &      & P     &  R    &  F1    \\
         \hline
        \multirow{2}{*}{Negation}    & dev  & 0.942 & 0.877 & 0.901  \\ 
                                     & test & 0.946 & 0.880 & 0.909  \\
        \hline
        \multirow{2}{*}{Uncertainty} & dev  & 0.959 & 0.953 & 0.956  \\
                                     & test & 0.948 & 0.961 & 0.954  \\
        \hline
    \end{tabular}
    \caption{Precision, recall, and F1 score of the negation and uncertainty classifiers on the SFU dev and test sets.}
    \label{tab:bow_eval}
\end{table}

\section{Additional Results on SFU+Amazon}
\label{app:amazon_results}

\subsection{Dataset Statistics}

\begin{table}[h!]
    \centering
    \small
    \begin{adjustbox}{center}
    \resizebox{0.5\textwidth}{!}{%
    \begin{tabular}{c|cccc}
        \multirow{2}{*}{Split} & \multirow{2}{*}{N} & Median & \multirow{2}{*}{\% Negated} & \multirow{2}{*}{\% Uncertain} \\
                               &                    & Length &                             &                               \\
        \hline
        Train & 109,889  &    12    &     22.7\%       &      17.9\%         \\
        Dev   & 6,631    &    12    &     19.6\%       &      15.2\%         \\
        Test  & 6,579    &    12    &     20.4\%       &      15.2\%         \\
        \hline
    \end{tabular}
    }
    \end{adjustbox}
    \caption{Statistics for the combined SFU and weakly labeled Amazon dataset.}
    \label{tab:data_sfu_amazon}
\end{table}

\vspace*{-1em}
\subsection{Example Reconstructions}

\begin{table}[h!]
    \centering
    \small
    \begin{adjustbox}{center}
    \resizebox{0.5\textwidth}{!}{%
    \begin{tabular}{|c|L|}
    \hline
        Input & {\em going home early.} \\
        \hline
              & {\em going home later.}\\ 
        Recon & {\em going home movies.}\\ 
              & {\em going home first.} \\ 
        \hline\hline
        Input & {\em this is a second outlet there is one on the dash.} \\ 
        \hline
              & {\em this is a second blender there is one on the floor.} \\
        Recon & {\em this is a second computer there is one on the seatbelt.} \\
              & {\em this is a second grease there is one on the cat.}  \\
        \hline\hline
        Input & {\em sometime it's just not enough volume. } \\
        \hline
              &{\em guess it's just not enough power.} \\
        Recon & {\em believe it works just not pleasant enough.} \\
              & {\em obviously it s just not enough control.}  \\
        \hline\hline
        Input & {\em I would never stay there again.}  \\
        \hline
              & {\em i would never stay them again.} \\
        Recon & {\em i would never rate that again.} \\
              & {\em i would not suggest that again.} \\
    \hline
    \end{tabular}
    }
    \end{adjustbox}
    \caption{Example reconstructions decoded by the INF+ADV+MI model from the SFU+Amazon test set.}
    \label{app:reconstructions_amazon}
\end{table}

\begin{table*}[h]
    \centering
    \resizebox{\textwidth}{!}{%
    \begin{tabular}{cc||cccc|cccc|cccc|cccc}
                              &        & \multicolumn{4}{c|}{+\INF}     & \multicolumn{4}{c|}{+\INF+\ADV}& \multicolumn{4}{c|}{+\INF+\MIN}& \multicolumn{4}{c}{+\INF+\ADV+\MIN}     \\
        Latent                & Factor &  MI   &  P    & R     & F1     &  MI   &  P    & R     & F1     &  MI   &  P    & R     & F1     &  MI    &  P    & R     & F1       \\
        \hline
        \multirow{2}{*}{$n$}  &  $n$   & 0.434 & 0.951 & 0.972 & 0.961  & 0.431 & 0.952 & 0.970 & 0.961  & 0.443 & 0.962 & 0.977 & 0.969  & 0.444 & 0.959 & 0.975 & 0.967    \\
                              &  $u$   & 0.013 & 0.551 & 0.584 & 0.544  & 0.012 & 0.551 & 0.581 & 0.547  & 0.007 & 0.537 & 0.560 & 0.527  & 0.007 & 0.543 & 0.569 & 0.537    \\
        \hline        
        \multirow{2}{*}{$u$}  &  $n$   & 0.013 & 0.555 & 0.579 & 0.550  & 0.008 & 0.553 & 0.573 & 0.550  & 0.008 & 0.543 & 0.561 & 0.535  & 0.007 & 0.548 & 0.560 & 0.548    \\
                              &  $u$   & 0.375 & 0.936 & 0.972 & 0.952  & 0.374 & 0.941 & 0.972 & 0.956  & 0.389 & 0.960 & 0.980 & 0.970  & 0.391 & 0.970 & 0.981 & 0.975    \\
        \hline      
        \multirow{2}{*}{$c$}  &  $n$   & 0.222 & 0.675 & 0.753 & 0.684  & 0.166 & 0.576 & 0.619 & 0.550  & 0.182 & 0.640 & 0.710 & 0.638  & 0.147 & 0.576 & 0.617 & 0.557    \\
                              &  $u$   & 0.198 & 0.643 & 0.746 & 0.649  & 0.144 & 0.567 & 0.626 & 0.538  & 0.169 & 0.638 & 0.740 & 0.641  & 0.136 & 0.574 & 0.637 & 0.551    \\
        \hline
    \end{tabular}}
    \caption{Mutual Information estimate between each latent and factor. Precision, recall, and F1 of the latent space classifiers for each factor. Shown are the mean values computed from 30 resamples of the latents for each example on the SFU+Amazon test set.}
    \label{tab:prediction_results_plus}
\end{table*}

\begin{table*}[ht!]
\small
    \centering
    \resizebox{\textwidth}{!}{%
    \begin{tabular}{c|cc|cc|cc|cc|cc}
                                \multicolumn{3}{c}{}           & \multicolumn{2}{c}{}         & \multicolumn{2}{c}{+\INF}    & \multicolumn{2}{c}{+\INF}    &  \multicolumn{2}{c}{+\INF+\ADV}     \\
        \multicolumn{1}{c}{}  & \multicolumn{2}{c}{\ELBO}      & \multicolumn{2}{c}{+\INF}    & \multicolumn{2}{c}{+\ADV}    & \multicolumn{2}{c}{+\MIN}    &  \multicolumn{2}{c}{+\MIN}         \\
                              & $u$        & $c$               & $u$        & $c$             & $u$        & $c$             & $u$        & $c$             & $u$       & $c$                   \\
        \hline
         \mrt{$n$}            & \mrt{0.706} & 0.008            & \mrt{0.200} & 0.002          & \mrt{0.193} & 0.007          & \mrt{0.129} & 0.008          & \mrt{0.159}     & 0.001           \\
                              &       & {\tiny ($\pm0.053$)}   &       & {\tiny ($\pm0.098$)} &       & {\tiny ($\pm0.076$)} &       & {\tiny ($\pm0.058$)} &           & {\tiny ($\pm0.043$)}  \\
        \hline
         \mrt{$u$}            & \mrt{-} & 0.001                & \mrt{-}     & 0.001          & \mrt{-}     & 0.010          & \mrt{-}     & 0.001          & \mrt{-}       & 0.005                 \\
                              &     & {\tiny ($\pm0.058$)}     &     & {\tiny ($\pm0.097$)}   &     & {\tiny ($\pm0.070$)}   &     & {\tiny ($\pm0.060$)}   &     & {\tiny ($\pm0.037$)}  \\
        \hline
    \end{tabular}
    }
    \caption{Pearson's correlation coefficients between each pair of latent representations across models on SFU+Amazon for each disentanglement objective. As the content space is 62-dimensional, we compute the correlation coefficient between each dimension and report the mean and standard deviation below in parentheses.}
    \label{tab:covariances_plus}
\end{table*}

\newpage
\subsection{Controlled Generation}
\label{app:control_results_amazon}

As discussed in \cref{sec:controlled_generation}, modifying the negation or certainty status of the input works well only in one direction (negated to positive, uncertain to certain). While the content space is uninformative of negation and uncertainty, we found that it \textit{is} highly informative of sentence length, which hinders the decoder from adding or removing tokens to satisfy the negation/uncertainty representations. Examples of successful and failed transfers illustrating this phenomenon are given in tables \ref{tab:successful_transfers_amzn} and \ref{tab:failed_transfers_amzn}. \cref{tab:sent_len_results_amzn} reports the $R^2$ of linear regressions of the number of tokens in the input on samples from the negation, uncertainty, and content distributions, respectively. 

Note that the transferred results have the same number of tokens as the inputs, and the decoder even repeats tokens where necessary to meet the length requirement (e.g., {\em fit fit} in the last negation example in \cref{tab:successful_transfers_amzn}). In general, the decoder seems to satisfy the value of the transferred factor by changing tokens in the input. This is clear in the last uncertainty example in \cref{tab:successful_transfers_amzn}, where the \textit{certain} input is correctly changed to \textit{uncertain}, but the uncertainty cue replaces the negation cue.

\begin{table}[h!]
    \centering
    \begin{tabular}{l|ccc}
        Latent      & train & dev   & test  \\
        \hline 
        Negation    & 0.048 & 0.057 & 0.063 \\
        Uncertainty & 0.051 & 0.034 & 0.041 \\
        Content     & 0.901 & 0.903 & 0.900 \\
        \hline
    \end{tabular}
    \caption{$R^2$ of the sentence length regression models on $z$'s sampled from INF+ADV+MIN model. $R^2 = 1$ represents perfect predictive ability.}
    \label{tab:sent_len_results_amzn}
\end{table}

\begin{table}[h!]
    \centering
    \small
    \begin{adjustbox}{center}
    \resizebox{0.5\textwidth}{!}{%
    \begin{tabular}{|c|L|}
    \hline
        \multirow{6}{*}{Neg} & i received my lodge grill griddle and it is extremely well made. \\
         & {\em i got my kitchen steel corkscrew but it is {\bf not} well made.}\\ 
        \cline{2-2}
         & if you do {\bf n't} have one get it now.\\
         & {\em if you must have need one used it now.} \\
        \cline{2-2}
         & but it does {\bf not} fit the tmo galaxy s. \\
         & {\em but it does fit fit the net galaxy s.} \\
        \hline
        \multirow{6}{*}{Unc} & it is light in weight and easy to clean. \\
         & {\em it is slid in sponges and {\bf seems} to clean.}\\ 
        \cline{2-2}
         & it snaps onto the phone in two pieces.\\
         & {\em it \textbf{should} affect the phone in two ways.} \\
        \cline{2-2}
         & but we did \textbf{n't} use it on the other one. \\
         & {\em but we {\bf would} have use it on the other one.} \\
        \hline
    \end{tabular}
    }
    \end{adjustbox}
    \caption{\textbf{Successful} inversions of negation and uncertainty. Inputs are in standard font and reconstructions are in {\em italics}. \textbf{bold} tokens indicate negation/uncertainty cues.}
    \label{tab:successful_transfers_amzn}
\end{table}

\begin{table}[h!]
    \centering
    \small
    \begin{adjustbox}{center}
    \resizebox{0.5\textwidth}{!}{%
    \begin{tabular}{|c|L|}
    \hline
        \multirow{6}{*}{Neg} & it is light in weight and easy to clean. \\
         & {\em it is tight in light and easy to clean.}\\ 
        \cline{2-2}
         & the x s were fine until i washed them.\\
         & {\em the iphone s worked fine and i return them.} \\
        \cline{2-2}
         & used for months and it is still going strong. \\
         & {\em used for now and it is still going strong.} \\
        \hline
        \multirow{6}{*}{Unc} & i received my lodge grill griddle and it is extremely well made. \\
         & {\em i received this roasting s cooking and it is still well made .}\\ 
        \cline{2-2}
         & glad i found these at a reasonable price.\\
         & {\em do i have these at a good price.} \\
        \cline{2-2}
         & it broke immediately when i put it on. \\
         & {\em it was immediately when i put it on.} \\
        \hline
    \end{tabular}
    }
    \end{adjustbox}
    \caption{\textbf{Failed} inversions of negation and uncertainty. Inputs are in standard font and reconstructions are in {\em italics}. \textbf{bold} tokens indicate negation/uncertainty cues.}
    \label{tab:failed_transfers_amzn}
\end{table}

\newpage
\section{Results on SFU+Yelp}
\label{app:yelp_results}

\begin{table}[h!]
    \centering
    \begin{adjustbox}{center}
    \resizebox{0.5\textwidth}{!}{%
    \begin{tabular}{c|cccc}
        \multirow{2}{*}{Split} & \multirow{2}{*}{N} & Median & \multirow{2}{*}{\% Negated} & \multirow{2}{*}{\% Uncertain} \\
                               &                    & Length &                       & \\
        \hline 
        Train & 390,409  &    9    &     14.5\%       &      8.8\%         \\
        Dev   & 5,971    &    9    &     15.5\%       &      9.1\%         \\
        Test  & 3,118    &    9    &     12.9\%       &      10.4\%        \\
        \hline
    \end{tabular}
    }
    \end{adjustbox}
    \caption{Statistics for the combined SFU and weakly labeled Yelp dataset.}
    \label{tab:data_sfu_yelp}
\end{table}

We here provide a complete evaluation of our models on a combination SFU+Yelp dataset, analogous to that performed on SFU+Amazon in the main text. Due to the shorter average length in tokens of the Yelp examples and the consequent reduction in compute power required, we were able to construct a combined dataset using the entire Yelp dataset with weak negation and uncertainty labels assigned according to the method described in \S~4.1.1 of the main text. Statistics of this dataset are given in \cref{tab:data_sfu_yelp}.

\begin{figure}[h!]
    \centering
    \includegraphics[scale=0.3]{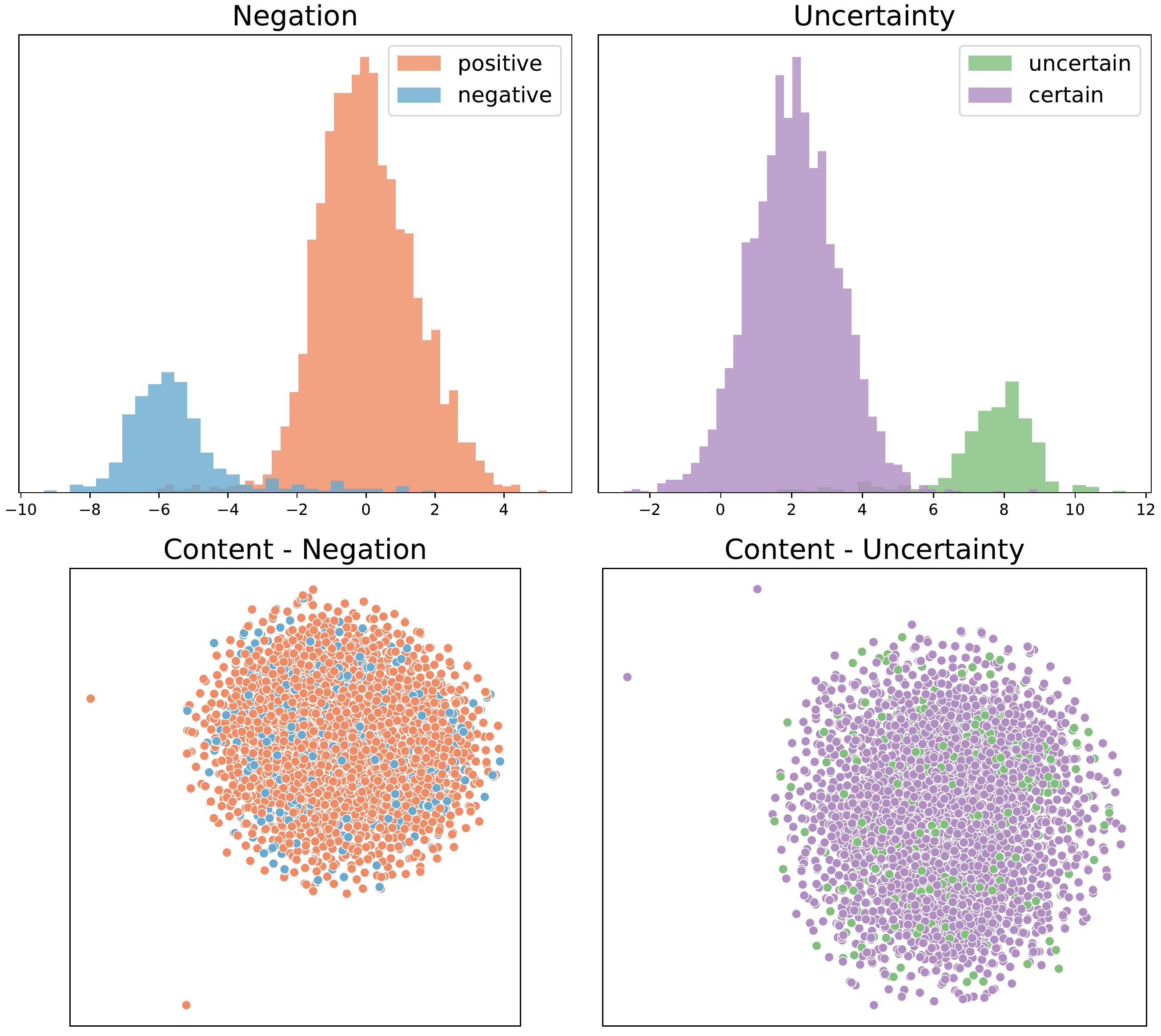}
    \caption{Histogram and t-SNE plots of the negation, uncertainty, and content spaces learned by the \ELBO+\INF+\ADV+\MIN~ model on SFU+Yelp test set.}
    \label{fig:latent_viz_yelp}
\end{figure}

First, visualizations of the latent distributions in \cref{fig:latent_viz_yelp} show 1) that the negation and uncertainty spaces are bimodal and discriminative of their corresponding factors while the content space is discriminative of neither; 2) the latent spaces are smooth and approximately normally distributed, despite two outliers.

\subsection{Disentanglement}

The mean predictive performance of each latent space for each objective is given in \cref{tab:prediction_results_yelp}, computed from 30 resamples of the latents for each test example. We also report the mean MI between each latent space and each factor over 30 resamples, as this provides additional insight into the informativeness of each space. Like the results on SFU+Amazon, these results show that the negation and uncertainty space are highly informative of their target factors and uninformative of the non-target factors. Additionally, the content space is informative of neither, showing near-random prediction performance. However, unlike the SFU+Amazon results, all objectives perform approximately equally, although the full \ELBO+\INF+\ADV+\MIN~ objective does reduce the informativeness of the content space slightly compared to the other objectives.

\begin{table*}[p]
\resizebox{\textwidth}{!}{%
    \centering
    \begin{tabular}{cc||cccc|cccc|cccc|cccc|cccc}
                              &        & \multicolumn{4}{c|}{\ELBO}     & \multicolumn{4}{c|}{+\INF}     & \multicolumn{4}{c|}{+\INF+\ADV}& \multicolumn{4}{c|}{+\INF+\MIN}&  \multicolumn{4}{c}{+\INF+\ADV+\MIN}     \\
        Latent                & Factor &  MI   &  P    & R     & F1     &  MI   &  P    & R     & F1     &  MI   &  P    & R     & F1     &  MI   &  P    & R     & F1     &  MI    &  P    & R     & F1       \\
        \hline
        \multirow{2}{*}{$n$}  &  $n$   & 0.020 & 0.560 & 0.628 & 0.555  & 0.308 & 0.916 & 0.954 & 0.934  & 0.307 & 0.913 & 0.954 & 0.932  & 0.308 & 0.911 & 0.953 & 0.931  & 0.307  & 0.920 & 0.954 & 0.936    \\
                              &  $u$   & 0.007 & 0.530 & 0.558 & 0.512  & 0.012 & 0.543 & 0.596 & 0.527  & 0.012 & 0.560 & 0.628 & 0.556  & 0.010 & 0.550 & 0.612 & 0.538  & 0.012  & 0.557 & 0.623 & 0.548    \\
        \hline        
        \multirow{2}{*}{$u$}  &  $n$   & 0.017 & 0.549 & 0.627 & 0.505  & 0.013 & 0.540 & 0.577 & 0.525  & 0.017 & 0.543 & 0.584 & 0.527  & 0.012 & 0.542 & 0.582 & 0.525  & 0.016  & 0.543 & 0.582 & 0.530    \\
                              &  $u$   & 0.036 & 0.583 & 0.674 & 0.560  & 0.281 & 0.926 & 0.963 & 0.943  & 0.282 & 0.920 & 0.962 & 0.940  & 0.282 & 0.922 & 0.963 & 0.941  & 0.284  & 0.921 & 0.962 & 0.940    \\
        \hline      
        \multirow{2}{*}{$c$}  &  $n$   & 0.236 & 0.654 & 0.783 & 0.671  & 0.183 & 0.601 & 0.705 & 0.591  & 0.173 & 0.575 & 0.658 & 0.553  & 0.174 & 0.575 & 0.656 & 0.551  & 0.162  & 0.561 & 0.630 & 0.531    \\
                              &  $u$   & 0.234 & 0.617 & 0.761 & 0.624  & 0.197 & 0.591 & 0.716 & 0.579  & 0.211 & 0.584 & 0.702 & 0.569  & 0.180 & 0.577 & 0.689 & 0.557  & 0.196  & 0.577 & 0.684 & 0.559    \\
        \hline
    \end{tabular}}
    \caption{Mutual Information estimate between each latent and factor. Precision, recall, and F1 of the latent space classifiers for each factor. Shown are the mean values computed from 30 resamples of the latents for each example on the SFU+Yelp test set.}
    \label{tab:prediction_results_yelp}
\end{table*}

\begin{figure*}[p]
    \centering
    \includegraphics[width=0.95\textwidth]{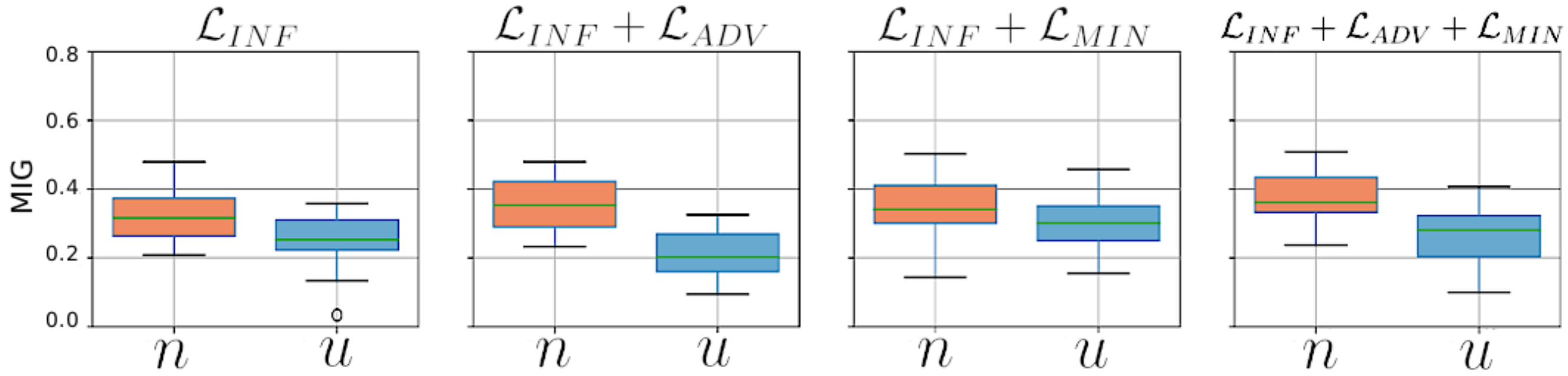}
    \caption{Box plots of the Mutual Information Gap (MIG) for each disentanglement objective for the negation and uncertainty factors computed on the Yelp test set.}
    \label{fig:migs_mis_yelp}
\end{figure*}

Box plots of the MIG and corresponding MI values for each disentanglement objective are given in \cref{fig:migs_mis_yelp}. In general, there is less disentanglement on SFU+Yelp than on SFU+Amazon (MIG $\approx$ 0.4 vs MIG $\approx$ 0.6 on SFU+Amazon). A comparison of the MI values reported in \cref{tab:prediction_results_yelp} to those in Table 1 of the main text shows the cause: the negation and uncertainty latents in SFU+Yelp are less predictive of their respective target factors (MIs $\approx$ 0.3 vs $\approx$ 0.4 on SFU+Amazon) while the content space is more predictive of these factors (e.g., content-uncertainty MI = 0.196 on SFU+Yelp vs 0.136 on SFU+Amazon). This may be due to a data mismatch between SFU and Yelp, since SFU is a dataset of product reviews, while Yelp contains mostly restaurant and store reviews.

\begin{table*}[p]
\small
    \centering
    \begin{tabular}{lc|ccc|ccc}
        \multicolumn{2}{c}{} & \multicolumn{3}{c}{}     & \multicolumn{3}{c}{+\INF+\ADV}  \\
        \multicolumn{2}{c}{} & \multicolumn{3}{c}{+\INF}& \multicolumn{3}{c}{+\MIN}  \\
        Factor       &  Pass & P       & R      & F1    & P       & R      & F1    \\
        \hline
\multirow{2}{*}{Neg} & 1     & 0.962   & 0.939  & 0.950 & 0.960   & 0.942  & 0.951 \\
                     & 2     & 0.932   & 0.887  & 0.908 & 0.944   & 0.907  & 0.924 \\
        \hline
\multirow{2}{*}{Unc} & 1     & 0.975   & 0.955  & 0.965 & 0.982   & 0.950  & 0.965 \\
                     & 2     & 0.898   & 0.900  & 0.899 & 0.933   & 0.894  & 0.912 \\
        \hline
    \end{tabular}
    \caption{Consistency of the decoder with the ground-truth values of negation and uncertainty evaluated on the SFU+Yelp test set. Pass 1 refers to the predictions from the original inputs. Pass 2 refers to the predictions from the re-encoded reconstructions. Pass 1 can be considered an upper bound on the performance of pass 2.}
    \label{tab:consistency_results_yelp}
\end{table*}

\begin{table*}[p]
\small
    \centering
    \begin{adjustbox}{center}
    \begin{tabular}{c|cc|cc|cc|cc|cc}
                                \multicolumn{3}{c}{}           & \multicolumn{2}{c}{}         & \multicolumn{2}{c}{+\INF}    & \multicolumn{2}{c}{+\INF}    &  \multicolumn{2}{c}{+\INF+\ADV}     \\
        \multicolumn{1}{c}{}  & \multicolumn{2}{c}{\ELBO}      & \multicolumn{2}{c}{+\INF}    & \multicolumn{2}{c}{+\ADV}    & \multicolumn{2}{c}{+\MIN}    &  \multicolumn{2}{c}{+\MIN}         \\
                              & $u$        & $c$               & $u$        & $c$             & $u$        & $c$             & $u$        & $c$             & $u$       & $c$                   \\
        \hline
         \mrt{$n$}            & \mrt{0.090} & 0.003            & \mrt{0.135} & 0.001          & \mrt{0.164} & 0.012          & \mrt{0.110} & 0.006          & \mrt{0.123}     & 0.002           \\
                              &       & {\tiny ($\pm0.013$)}   &       & {\tiny ($\pm0.047$)} &       & {\tiny ($\pm0.038$)} &       & {\tiny ($\pm0.034$)} &           & {\tiny ($\pm0.028$)}  \\
        \hline
         \mrt{$u$}            & \mrt{-} & 0.001                & \mrt{-}     & 0.003          & \mrt{-}     & 0.003          & \mrt{-}     & 0.004          & \mrt{-}       & 0.010                 \\
                              &     & {\tiny ($\pm0.023$)}     &     & {\tiny ($\pm0.042$)}   &     & {\tiny ($\pm0.047$)}   &     & {\tiny ($\pm0.040$)}   &     & {\tiny ($\pm0.036$)}  \\
        \hline
    \end{tabular}
    \end{adjustbox}
    \caption{Pearson's correlation coefficients between each pair of latent representations across models on SFU+Yelp. As the content space is 62-dimensional, we compute the correlation coefficient between each dimension and report the mean and standard deviation.}
    \label{tab:covariances_yelp}
\end{table*}

\begin{table*}[p]
\small
    \centering
    \begin{tabular}{cl|ccccc}
                              &       &       &       &       &       & +\INF    \\
                              &       &       &       & +\INF & +\INF & +\ADV    \\
                              &       & \ELBO & +\INF & +\ADV & +\MIN & +\MIN    \\
        \hline
        \multirow{3}{*}{BLEU} & Train & 0.805 & 0.786 & 0.793 & 0.786 &  0.796   \\
                              & Dev   & 0.492 & 0.382 & 0.394 & 0.383 &  0.398   \\
                              & Test  & 0.400 & 0.298 & 0.309 & 0.300 &  0.315   \\
        \hline\hline
        \multirow{3}{*}{PPL}  & Train & 52.7  & 54.8  & 53.3  & 54.1  & 53.7      \\
                              & Dev   & 75.9  & 76.9  & 76.2  & 77.4  & 76.5      \\
                              & Test  & 106.2 & 105.9 & 106.0 & 106.7 & 105.5     \\
        \hline
    \end{tabular}
    \caption{Reconstruction self-BLEU and reconstruction perplexity (PPL) on each SFU+Yelp data split. Perplexity is computed using GPT-2.}
    \label{tab:generation_results_yelp}
\end{table*}

\subsection{Example Reconstructions}

\begin{table}[H]
    \centering
    \small
    \begin{adjustbox}{center}
    \resizebox{0.5\textwidth}{!}{%
    \begin{tabular}{|c|L|}
    \hline
        Input & {\em my jack and coke was seriously lacking.} \\
        \hline
              & {\em top brie and lobster was seriously lacking.}\\ 
        Recon & {\em my pastrami and swiss was totally lacking.}\\ 
              & {\em the camarones and coke was totally lacking.} \\ 
        \hline\hline
        Input & {\em plus the dude didn't even know how to work the computer.} \\ 
        \hline
              & {\em unfortunately the managers didn't do too enough to compliment the computer.} \\
        Recon & {\em unfortunately the women didn't even know how to honor the experience.} \\
              & {\em plus the baristas didn't even know how to control the language.}  \\
        \hline\hline
        Input & {\em the service was great and would gladly go back.} \\
        \hline
              & {\em the service was great and would easily go back.} \\
        Recon & {\em the service was exceptional and would probably go back.} \\
              & {\em the service was great and would certainly go back.} \\
        \hline\hline
        Input & {\em she could not and would not explain herself.} \\
        \hline
              & {\em she could not and would not explain herself.} \\
        Recon & {\em she could not and would not respond herself.} \\
              & {\em she could apologize and would not introduce herself.} \\
    \hline
    \end{tabular}
    }
    \end{adjustbox}
    \caption{Example reconstructions decoded by the INF+ADV+MI model from the SFU+Yelp test set.}
    \label{tab:reconstructions_yelp}
\end{table}

\subsection{Controlled Generation}

\cref{tab:attr_trans_yelp} reports the accuracies of attribute transfer on the SFU+Yelp test set. As reported for SFU+Amazon above, attribute transfer works well only when it is not necessary for the model to introduce additional tokens. 

\begin{table}[h!]
    \centering
    \begin{tabular}{cc}
        Transfer direction    & Accuracy \\
        \hline
        pos $\rightarrow$ neg & 0.38  \\
        neg $\rightarrow$ pos & 0.87  \\
        \hline
        cer $\rightarrow$ unc & 0.36  \\
        unc $\rightarrow$ cer & 0.86  \\
    \end{tabular}
    \caption{Attribute inversion accuracies by factor and direction of transfer using the \INF+\ADV+\MIN ~model on the SFU+Yelp test set.}
    \label{tab:attr_trans_yelp}
\end{table}

\begin{table}[h!]
    \centering
    \small
    \begin{adjustbox}{center}
    \resizebox{0.5\textwidth}{!}{%
    \begin{tabular}{|c|L|}
    \hline
        \multirow{6}{*}{Neg} & i totally agree but the way he said it was very arrogant. \\
         & {\em i even complained but the way but said it was \textbf{not} helpful.}\\ 
        \cline{2-2}
         & i will definitely come back for that and the singapore noodles.\\
         & {\em i absolutely \textbf{never} come back for that i ordered singapore noodles.} \\
        \cline{2-2}
         & overall i was \textbf{not} impressed and regret going. \\
         & {\em overall i was very impressed and recommended going.} \\
        \hline
        \multirow{6}{*}{Unc} & we \textbf{could}n't wait till he was gone. \\
         & {\em we don't wait till he was gone.}\\ 
        \cline{2-2}
         & room was very adequate quiet and clean.\\
         & {\em room \textbf{would} \textbf{seemed} quiet quiet and sanitary.} \\
        \cline{2-2}
         & the ending is as it \textbf{should} be. \\
         & {\em the ending is as it must be.} \\
        \hline
    \end{tabular}
    }
    \end{adjustbox}
    \caption{\textbf{Successful} inversions of negation and uncertainty on the SFU+Yelp test set. Inputs are in standard font and reconstructions are in {\em italics}. \textbf{bold} tokens indicate negation/uncertainty cues.}
    \label{tab:successful_transfers_yelp}
\end{table}

\section{Implementation Details}
\label{app:implementation}

We implement our model in PyTorch \citep{paszke2017automatic}. The encoder is a BiLSTM with 2 hidden layers and hidden size 256. The decoder is a 2 layer LSTM with hidden size 256. Embeddings for both the encoder and decoder are of size 256 and are randomly initialized and learned during training. The encoder and decoder use both word dropout and hidden layer dropout between LSTM layers, with a dropout probability of 0.5. We also use teacher forcing when training the decoder, with a probability of using the true previous token set to 0.5. We trained each model for 20 epochs with a batch size of 128 and the ADAM optimizer \citep{Kingma2015AdamAM} with a learning rate of $3\cdot 10^{-4}$.  Training took around 6.5 hours for each model on one Tesla v100 with 16GB of VRAM.

The latent space classifiers for the INF objective and the adversarial classifiers for the ADV objective both used a single linear layer with sigmoid activation. The adversarial classifiers were trained with a separate ADAM optimizer with learning rate of $3\cdot 10^{-4}$. For MI estimation as part of the MIN objective we used the code released alongside the CLUB paper \citep{Cheng_Hao_Dai_Liu_Gan_Carin_2020} \footnote{\url{https://github.com/Linear95/CLUB}}. For training the approximation network, we again use an ADAM optimizer with learning rate $5\cdot 10^{-4}$.

All hyperparameter weights were tuned by hand. We weight the KL divergence term of each latent space separately as follows: $\beta_{c}=0.01, ~\beta_{n} = \beta_{u} = 0.005$. We experimented with both higher KL weights and KL annealing schedules, but found that they did not improve disentanglement and higher weights tended to negatively impact reconstruction ability. For the individual objectives, we set the following weights: $\lambda_{INF} = 1.0, ~\lambda_{ADV} = 1.0, ~\lambda_{MIN} = 0.01$. While we found the model relatively robust to different values of $\lambda_{INF}$ and $\lambda_{ADV}$, the MIN objective was found to be quite sensitive to even small changes of both $\lambda_{MIN}$ and the learning rate of the MI approximation network.

\subsection{Mutual Information Gap}
\label{app:mig}

The Mutual Information Gap (MIG) is the difference in MI between the top-2 latent variables $\ell_{\{i,j\}}$ with respect to a given generative factor $k$, normalized to lie in $[0, 1]$, with higher values indicating a greater degree of disentanglement.

\begin{equation}
    \text{MIG}_{k} = \frac{1}{H[k]} \left( I({\bf z}^{(\ell_i)};k) - \max_{j \ne i} ~I({\bf z}^{(\ell_j)}; k)  \right)
\end{equation}

\noindent
where $\ell_i = \text{argmax}_{i'}~I({\bf z}^{(\ell_{i'})};k)$ is the latent space with the highest MI with the generative factor. We estimate the MI between latent representations and labels using the method proposed by \citet{Ross_2014}, implemeted using {\tt mutual\_info\_classif} in scikit-learn \cite{scikit-learn} using 30 resamples from the predicted latent distributions for each example.

\section{Ethical Considerations}
This work is foundational NLP research on semantics, and as such we do not foresee any immediate risks, ethical or otherwise. However, representation learning may be used as part of many downstream NLP tasks such as information extraction, classification, and natural language generation, which might be used for harmful surveillance, discrimination, and misinformation. 

The SFU, Amazon, and Yelp datasets used in this work do not attach unique identifiers such as user IDs or names to data instances such that the original authors might be identifiable. A manual review of a small random subset of the data did not reveal any overtly identifiable or harmful information within the text.

\end{document}